\documentclass[letterpaper, 10 pt, conference]{ieeeconf}  % Comment this line out if you need a4paper

\IEEEoverridecommandlockouts                              % This command is only needed if 
\usepackage{epsfig} % for postscript graphics files
\usepackage{amsmath} % assumes amsmath package installed
\usepackage{amssymb}  % assumes amsmath package installed
\usepackage{mathtools}
\usepackage{amsfonts}
\usepackage{cleveref}
\usepackage{makecell}
\usepackage{wrapfig}
\usepackage{subcaption}
\usepackage[table]{xcolor}
\usepackage{tabularx}
\usepackage{makecell}
\usepackage{multirow}

\definecolor{lightgray}{gray}{0.9}

\usepackage{algorithm}
\usepackage{algorithmic}
\usepackage{cite}

\title{\LARGE \bf
Reinforcement Learning Within the Classical Robotics Stack: A Case Study in Robot Soccer
}

\author{
Adam Labiosa\textsuperscript{1}\textsuperscript{*}, 
Zhihan Wang\textsuperscript{2}\textsuperscript{*}, 
Siddhant Agarwal\textsuperscript{2}, 
William Cong\textsuperscript{1}, 
Geethika Hemkumar\textsuperscript{2},\\ 
Abhinav Narayan Harish\textsuperscript{1},
Benjamin Hong\textsuperscript{1}, 
Josh Kelle\textsuperscript{2}, 
Chen Li\textsuperscript{1}, 
Yuhao Li\textsuperscript{1}, 
Zisen Shao\textsuperscript{1},\\ 
Peter Stone\textsuperscript{2,3}\textsuperscript{\textdagger},
Josiah P. Hanna\textsuperscript{1}\textsuperscript{\textdagger}% <-this % stops a space
\thanks{%
\begin{minipage}[t]{\linewidth}
\textsuperscript{1}University of Wisconsin--Madison. %\newline
\textsuperscript{2}The University of Texas at Austin. 
\textsuperscript{3}Sony AI.
\textsuperscript{*}Indicates equal contribution. %\newline
\textsuperscript{\textdagger}Indicates equal advising. \newline
All other authors listed in alphabetical order. \newline
{\tt\small Correspondence to: labiosa@wisc.edu}
\end{minipage}}%
}

\begin{document}

\maketitle
\thispagestyle{empty}
\pagestyle{empty}

%%%%%%%%%%%%%%%%%%%%%%%%%%%%%%%%%%%%%%%%%%%%%%%%%%%%%%%%%%%%%%%%%%%%%%%%%%%%%%%%
\begin{abstract}

% 1.	Introduction: Briefly introduce the problem or gap your research addresses.
% 2.	Method: Summarize your approach or methodology.
% 3.	Results: Highlight the main findings or results.
% 4.	Conclusion/Implications: State the significance or potential impact of your results.

Robot decision-making in partially observable, real-time, dynamic, and multi-agent environments remains a difficult and unsolved challenge.
Model-free reinforcement learning (RL) is a promising approach to learning decision-making in such domains, however, end-to-end RL in complex environments is often intractable. 
To address this challenge in the RoboCup Standard Platform League (SPL) domain, we developed a novel architecture integrating RL within a classical robotics stack, while employing a multi-fidelity sim2real approach and decomposing behavior into learned sub-behaviors with heuristic selection.
Our architecture led to victory in the 2024 RoboCup SPL Challenge Shield Division.
In this work, we fully describe our system's architecture and empirically analyze key design decisions that contributed to its success. 
Our approach demonstrates how RL-based behaviors can be integrated into complete robot behavior architectures.

\end{abstract}

%%%%%%%%%%%%%%%%%%%%%%%%%%%%%%%%%%%%%%%%%%%%%%%%%%%%%%%%%%%%%%%%%%%%%%%%%%%%%%%%
\section{INTRODUCTION} 

In the field of robotics, reinforcement learning (RL) has enabled complex and impressive behaviors \cite{akkaya2019solving, li2021reinforcement, dambrosio2024achievinghumanlevelcompetitive}. Despite the exciting advances in RL, the training and deployment of RL for strategic decision-making on physical robots in partially observable, real-time, dynamic, and multi-agent environments remains a challenge.

One particular domain that exhibits these challenges is the RoboCup Standard Platform League (SPL) \cite{nardi2014robocup}.
The SPL is part of the RoboCup initiative, which has driven advances in robotics over the past three decades \cite{kitano1997robocup}. In the SPL, teams of 5 or 7 humanoid NAO robots compete in soccer games. 
Each robot must be fully autonomous and act in real-time; and the presence of teammates and adversaries makes the domain highly dynamic.
In addition, it is a competitive environment that requires teams to quickly adapt to different opponents and improve their strategy between and within matches.
Teams participating in the SPL typically rely on a classical robot behavior architecture with complex hand-coded behaviors, and RL has had little use at the behavior level.

Toward the use of RL in partially observable, real-time, dynamic, and multi-agent environments, we introduce an RL-based robot architecture and training framework that we evaluate in the RoboCup SPL domain.
Using this architecture, our joint team across two universities, WisTex United, participated in and won the 2024 RoboCup SPL Challenge Shield Division. Over 8 games we won 7 and outscored opponents 39-7. 
To the best of our knowledge, our system represents the first successful use case of RL for high-level decision-making in the SPL domain. 
While specific to the SPL competition, our system design provides insights for roboticists seeking to apply RL in domains of similar complexity.

Our architecture is based upon a fairly standard classical robotics stack that decomposes perception, state estimation, behavior, and control into separate modules. 
Our main contributions are then to enable the use of RL as a central part of the behavior module that controls each robot's high-level, strategic decision-making.
The architecture enjoys the robustness of a modular approach, uses separately trained RL policies to achieve flexibility and versatility, and allows for improvement at deployment time.

To effectively train behaviors, we adopt a sim2real approach and use simulators of different fidelities. A lower fidelity simulator enables extensive full field training, whereas a higher fidelity simulator enables the robot to learn more precise ball control in critical situations. Furthermore, instead of training a monolithic policy for all game scenarios, we decompose the overall behavior into four learned sub-behaviors with different action and observation spaces. During games, we heuristically select between behaviors to integrate human knowledge into our strategy and enable rapid adjustment.

In this paper, we fully describe the key components of our architecture and training framework and then empirically study the importance of key design decisions.
Specifically, the main contributions of our work are:
\begin{itemize}
    \item We detail our novel RL-based robot behavior architecture and training framework that led to winning the RoboCup SPL Challenge Shield Division.
    \item We identify and describe key design choices in the architecture: multifidelity RL training, behavior decomposition into sub-behaviors, heuristic selection of sub-behaviors during deployment, and usage of different action and observation spaces across sub-behaviors.
    \item We analyze our key design choices in a series of ablation experiments. Our experiments validate the effectiveness of key aspects of our architecture, complementing our victory in the 2024 SPL Challenge Shield Division.
\end{itemize}

\section{BACKGROUND}

In this section, we provide background on reinforcement learning and describe related work on enabling RL in robotics and other use-cases of RL to target similar domains.

\subsection{Reinforcement Learning}

Reinforcement learning algorithms enable an agent to learn optimal actions in sequential decision-making environments.
% shorter, POMDP_v2
We formalize this environment as a Partially Observable Markov Decision Process (POMDP) $(\mathcal{S}, \mathcal{A}, \mathcal{P}, \mathcal{R}, \mathcal{O}, \Omega, \gamma)$, where $\mathcal{S}$ is the state space, $\mathcal{A}$ is the action space, $\mathcal{P}: \mathcal{S} \times \mathcal{A} \rightarrow \Delta(\mathcal{S})$ is the transition function, $\mathcal{R}: \mathcal{S} \times \mathcal{A} \rightarrow \mathbb{R}$ is the reward function, $\mathcal{O}$ is the observation space, $\Omega: \mathcal{S} \times \mathcal{A} \rightarrow \Delta(\mathcal{O})$ is the observation model, and $\gamma$ is the discount factor.
In a POMDP, the agent takes in the history of observations or a belief state and outputs an action. The objective is to maximize the expected cumulative reward, defined as $J(\pi) \coloneqq \mathbf{E}[\sum_{t=0}^\infty \gamma^t \mathcal{R}(s_t,a_t)]$. It should be noted that even though we are interested in the multi-robot SPL domain, from the point of view of any single robot, the actions of other robots are represented as just part of the state transition function.
%
% It should be noted that even though we are interested in the multi-robot SPL domain, from the point of view of any single robot, the actions of other robots are represented as just part of the state transition function.
%
% In this POMDP, the agent does not directly observe the true state but instead receives estimated observations. The agent’s policy is based on a limited history of previous observations, with different policies potentially using different lengths of history. The objective is to maximize the expected cumulative reward.

% shorter, and POMDP
% We formalize this environment as a Partially Observable Markov Decision Process (POMDP) $(\mathcal{S}, \mathcal{A}, \mathcal{P}, \mathcal{R}, \mathcal{O}, \Omega)$, where $\mathcal{S}$ is the state space, $\mathcal{A}$ is the action space, $\mathcal{P}: \mathcal{S} \times \mathcal{A} \rightarrow \Delta(\mathcal{S})$ is the transition function, $\mathcal{R}: \mathcal{S} \times \mathcal{A} \rightarrow \mathbb{R}$ is the reward function, $\mathcal{O}$ is the observation space, and $\Omega: \mathcal{S} \times \mathcal{A} \rightarrow \Delta(\mathcal{O})$ is the observation model. In a POMDP, the agent does not directly observe the state but receives observations, so its policy is defined as $\pi: \mathcal{O} \rightarrow \Delta(\mathcal{A})$, mapping observations to action distributions. The objective is to maximize the expected cumulative reward $J(\pi) \coloneqq \mathbf{E}[\sum_{t=0}^\infty \mathcal{R}(s_t,a_t)]$, where $a_t \sim \pi(o_t)$ and $o_t \sim \Omega(s_t, a_{t-1})$. 

\subsection{Related Work}

In this section we discuss related work in RL for robotics, multi-fidelity simulation, and robot soccer.

\subsubsection{Reinforcement Learning in Robotics}

Reinforcement Learning (RL) has significantly advanced robot learning \cite{tang2024deep}. Specifically, the paradigm of sim2real transfer has shown success on a body of work on locomotion for bipedal robots \cite{siekmann2020learning, castillo2022reinforcement, duan2024learning, li2021reinforcement, beranek2021behavior, kouppas2021hybrid, qin2024heuristics, li2019using}; however, these works do not use RL for high-level learning or in a domain as challenging as the SPL robot soccer league.

Other research has explored high-level decision-making with hierarchical approaches \cite{nachum2019multi, li2020learning, li2021planning}, but these works did not deal with bipedal robots and studied domains with stable and predictable dynamics unlike in the SPL. Some works have investigated training exclusively high-level behaviors in abstract simulations \cite{truong2023rethinking, zhang2024back}, but they also do not address many of the SPL domain complexities.
Our work also distinguishes itself by manually decomposing behaviors rather than training a single high-level policy. This approach allows for more fine-grained control and potentially better transferability to real-world scenarios.

\subsubsection{Multi-fidelity Simulation}

Our approach utilizes two levels of simulation fidelity. Many works use multi-fidelity simulation with RL to maximize sample efficiency and policy performance \cite{bhola2023multi, cutler2014reinforcement, cutler2015real, khairy2024multi, beard2022black}, however, most do not apply the approach to physical robots. A few works have applied multi-fidelity simulation to sim2real transfer \cite{suryan2020multifidelity, ryou2024multi} but they have trained a single policy through increasing levels of realism. In contrast, our work focuses on training multiple decomposed policies across different fidelities.

\subsubsection{Robot Soccer}

Within the domain of robot soccer \cite{kitano1997robocup, hong2021ai, smit2023scaling, antonioni2021game}, numerous studies have applied RL techniques. Many of these works are conducted in simulation environments \cite{AB05,abreu2023designingskilledsoccerteam, huang2021tikickplayingmultiagentfootball, lin2023tizeromasteringmultiagentfootball, liu2022motor, liu2019emergent}, as opposed to physical robots. Others focus on wheeled or quadrupedal robots \cite{da2021deep, merke2002karlsruhe, riedmiller2009reinforcement, huang2023creating, ji2023dribblebot, ji2022hierarchical} rather than a bipedal system.

Haarnoja et al. \cite{haarnoja2024learning} learn joint movements directly such that a bipedal robot was able to learn a policy that demonstrates strong performance 1 vs 1 robot soccer. This work is limited with the use of a global motion capture system to provide precise state estimates and therefore does not solve many of the challenges present in the SPL. Heuristics have been explored for teamwork in the robot soccer domain \cite{ros2009case}, but have not been combined with RL policies.
    
\section{RoboCup Standard Platform League: Domain Challenges and Reinforcement Learning Integration}

In this section, we describe the robotics challenges raised by the SPL and then describe the additional challenges that must be overcome to develop an RL-based architecture for the domain.

\subsection{RoboCup Standard Platform League}

The SPL presents a challenging robotics task for numerous reasons. First, all of the robots must be fully autonomous, with all perception and the control onboard the robot.
Second, sparse wireless communication is available but limited by competition rules, delayed, and unreliable at a competition venue.
Third, the domain requires real-time perception from visual and proprioceptive data processed on a CPU so the robots operate with significant uncertainty about the full state of the world.
Fourth, the domain is highly-dynamic because the positions of the robots and ball are constantly changing.
Fifth, effective team behavior requires each robot to coordinate to fill the right role at the right time under these dynamic match conditions.
Last, robots must react to unpredictable opponent behaviors and balance assertiveness with penalty avoidance.

This combination of factors creates a challenging decision-making domain where robots must rapidly process incomplete information to formulate strategies.

The SPL consists of two divisions: the Champion's Cup Division features 7v7 games, and the Challenge Shield Division is for 5v5 games. While the former is generally more competitive, the Challenge Shield still serves as a strong baseline since all teams have access to code from previous years' top performers. As we describe below, we based our RL-based architecture on Team B-Human's publicly available architecture that was used to win the 2023 Champion's Cup \cite{BHumanCodeReleaseDocumentation2023}. Other teams in both divisions also built their approach upon code from B-Human.

\subsection{Challenges with Applying RL in the SPL Domain}

Despite RL demonstrating success in simulated 2D and 3D domains and showing promise for skills such as walking, no SPL team, to the best of our knowledge, has successfully used RL to develop the primary strategic decision-making of their robots.
In this section, we describe the challenges with applying RL in the SPL domain.

\subsubsection{Challenge of Using RL for End-to-End Learning}
Much robot RL research has focused on end-to-end learning where a single neural network controls a robot at the lowest level of control.
For instance, in the soccer domain, Tirumala et al. \cite{tirumala2024learning} showed that robots could be trained end-to-end to play short 1 vs. 1 matches from vision. While impressive, SPL games span 20 minutes and require multiple robots to coordinate under more complex rules. These factors make end-to-end RL learning for the SPL require prohibitively high compute resources.

\subsubsection{Challenges with Integrating RL for High-Level Decision Making in SPL}

Integrating RL into high-level decision-making for robot soccer presents several interconnected challenges. While the SPL community has access to refined low-level skills from previous teams, developing RL policies that effectively utilize these skills is difficult due to the sim2real gap and limitations in simulation technology. Specifically, the available high-fidelity simulator, though relatively accurate in modeling physics, is computationally inefficient and does not parallelize for GPU-training. As well, the competitive dynamics, complexity, and multi-agent interactions inherent in robot soccer make training a monolithic RL policy that handles all situations infeasible in the high-fidelity simulator.

% Our setup
\section{REINFORCEMENT LEARNING WITHIN A COMPLETE ROBOT SYSTEM}

In this section we describe our system and the key design decisions that contributed to winning the SPL Challenge Shield Division. 

\begin{table*}[ht]
\vspace{2mm}
    \centering
    \rowcolors{2}{lightgray}{}
    \begin{tabularx}{\linewidth}{p{2cm}p{3cm}XX}
    \hline
    Policy & Action Space & Action Space Description & Observation Space  \\
    \hline
    \textsc{Mid-field} & [$\Delta \Theta$] & 
    % Adjusts position to move toward to kick the ball in the direction of the angle output. 
    Adjusts the desired kicking angle relative to a global reference frame.
    % Adjusts goal position to walk toward. Angle output defines the desired kicking angle.
    Action is clipped for stability.
    % Adjust the robot's position to align with the calculated kicking angle.
    % The robot adjusts its position to kick the ball the angle. Action is clipped for stability. 
    & [Ball, Can kick 1-hot, Goal center, All goalposts, Field sides, Last 3 ball positions]\\
    \textsc{Ball Duel} & [$\Delta X$, $\Delta Y$, $\Delta \Theta$] & Controls the robot’s movement at an egocentric (robot-centered) velocity in the x, y directions, and adjusts its orientation (theta). & [Ball, Can kick 1-hot, Closest teammate to goal, All goalposts, Field sides, Last 3 ball positions]\\
    \textsc{Near-goal} & [$\Delta X$, $\Delta Y$, $\Delta \Theta$] & Same as \textsc{Ball Duel}. & [Ball, Opponent goalposts, Last 3 ball positions]\\
    \textsc{Positioning} & [$\Delta X$, $\Delta Y$, $\Delta \Theta$, $Stand$] & Similar to \textsc{Ball Duel} and \textsc{Near-goal} but with the addition of a stand thresholded action. & [Ball, Strategy position, All defenders, All goalposts, Field sides, Last 3 ball positions]  \\
    \hline
    \end{tabularx}
    \caption{Action and observation space details for each sub-policy.}
\label{tab:act_obs_spaces}
\end{table*}

\subsection{Robot Architecture Setup}
\label{subsec:arch_setup}

\begin{figure*}[ht]
\vspace{-3mm}
    \centering
    \includegraphics[width=0.8\linewidth]{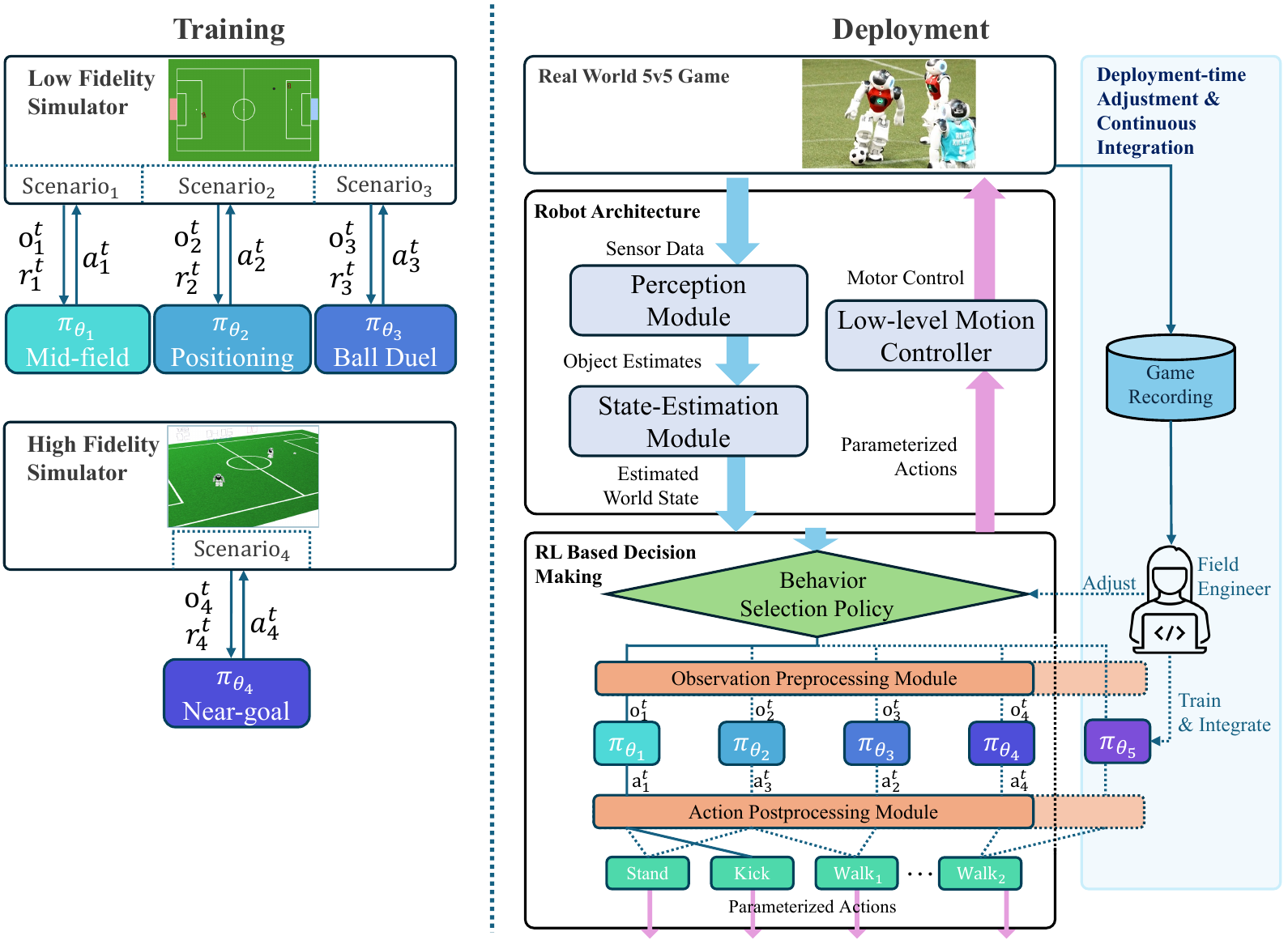}
    \caption{Architecture of our training and deployment system for robot soccer. The left side illustrates our training setup, utilizing both high-fidelity (SimRobot) and low-fidelity (AbstractSim) simulators to train policies with different action spaces under various scenarios. The right side depicts our deployment architecture for real-world 5v5 games, built upon the B-Human team's classical robotics framework. It includes a Perception Module processing sensor data, a State-Estimation Module computing robot and ball positioning, and our RL decision module. The RL module, receiving processed observations, uses a heuristic Behavior Selection Policy to choose appropriate sub-behavior policies, which determine actions executed by the low-level controller. Our heuristic approach allows for dynamic play style adjustments and easy integration of new policies, and facilitates continuous improvement at deployment time.}
    \label{fig:trainingdeployment}
    \vspace{-3mm}
\end{figure*}

We build our system architecture (\Cref{fig:trainingdeployment}) on top of an existing classical robotics framework. Specifically, we leverage the complete robot architecture developed by the B-Human team \cite{BHumanCodeReleaseDocumentation2023}, which includes finely tuned motion primitives, robot localization, and object perception modules. In doing so, we avoid the need to learn robot perception and locomotion and can avoid the computational expense of end-to-end RL. Instead, the RL policies we train take high-level aspects of the game as input (e.g., ball and robot positions) and output high-level controls (see \Cref{tab:act_obs_spaces}).

\subsection{Simulation Environments}

\begin{figure}[htbp]
    \centering
    \begin{minipage}[t]{0.48\linewidth} % Adjusted width slightly to accommodate thinner space
        \centering
        \includegraphics[width=\linewidth]{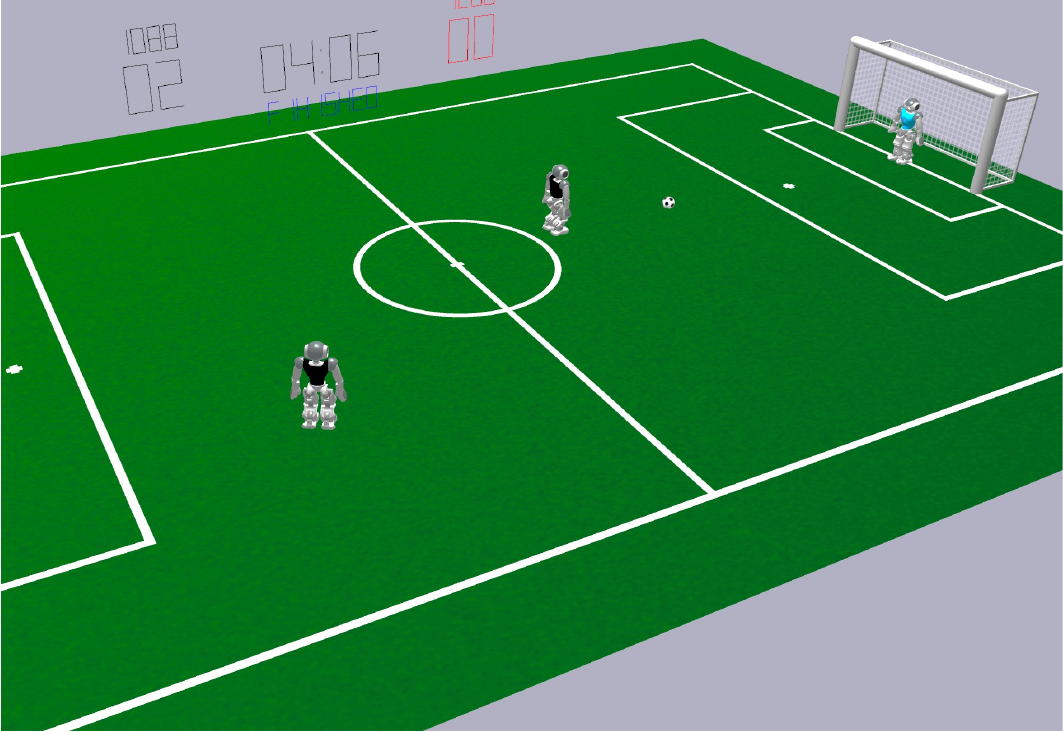}
        \caption{High-fidelity simulation SimRobot developed by the B-Human RoboCup Team. Physics are based on the Open Dynamics Engine.}
        \label{fig:simrobot}
    \end{minipage}
    \hspace{0.01\linewidth} % Reduced horizontal space to 0.01
    \begin{minipage}[t]{0.48\linewidth} % Adjusted width slightly to accommodate thinner space
        \centering
        \includegraphics[width=\linewidth]{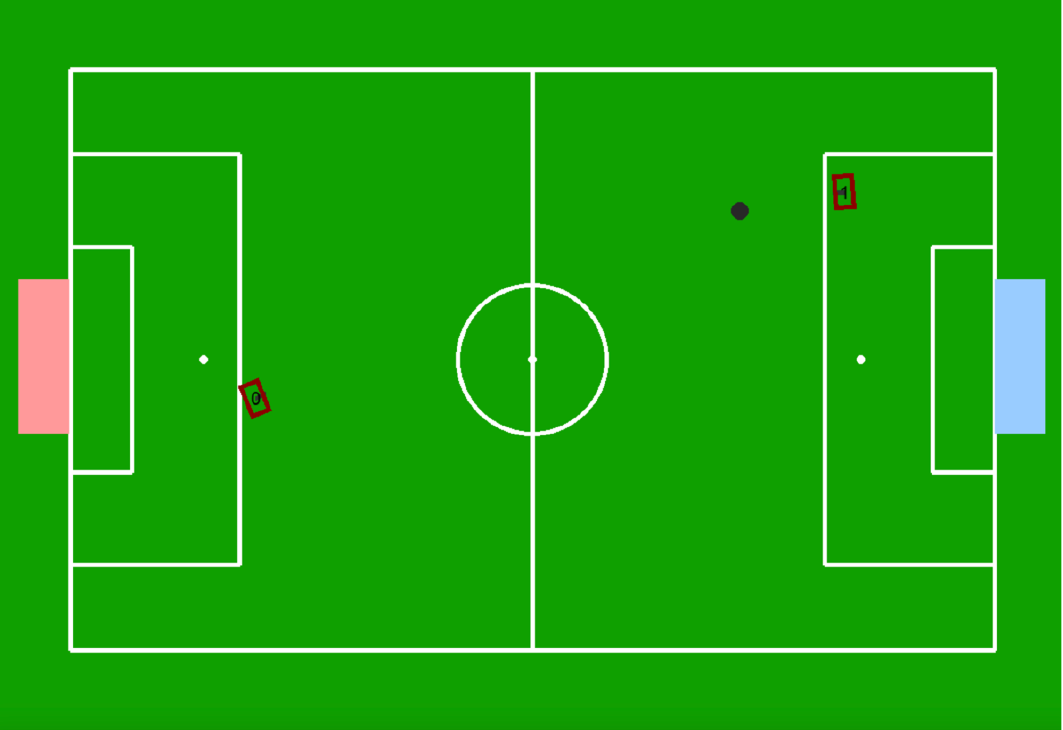}
        \caption{Custom low-fidelity simulation AbstractSim, in which robots are modeled as rectangles and joint movements are abstracted.}
        \label{fig:abstractsim}
    \end{minipage}

\vspace{-3mm}
\end{figure}

To enable RL-trained behaviors on physical robots, we adopt the sim2real paradigm -- train RL in simulation and deploy on physical robots. A challenge is that our high-fidelity simulator is prohibitively slow for RL training (\Cref{fig:simrobot}. To address this challenge, we observe that the full complexity and precision of high-fidelity simulation is unnecessary in most gameplay scenarios. Instead, the primary requirement is to choose the direction for a better position and thus a simplified simulation is generally sufficient for training.

Our low-fidelity simulation, AbstractSim, is a lightweight system that we developed to enable fast and efficient training (see \Cref{fig:abstractsim}). In AbstractSim, robots are represented as simple rectangles; their movement is modeled without considering the complexity of joints, legs, or feet; and the ball follows simple kinematic motion. This simulator drastically reduces the computational load, allowing us to efficiently train agents across the entire field.

For high-fidelity training, we use the SimRobot simulator (\Cref{fig:simrobot}), developed by the B-Human team. Training across the entire field in such a high-fidelity environment would be impractical, with runs taking weeks for a single robot. Consequently, we restrict high-fidelity training to critical near-goal scenarios where precision and fine-grained control are paramount. 

\subsection{Reinforcement Learning Behavior Decomposition}
\label{sec:rlbd}

Instead of training a monolithic policy for all game scenarios, we decompose the overall behavior into four sub-behaviors trained with PPO \cite{schulman2017proximal, stable-baselines3} which make use of different simulator fidelities, and action and observation spaces (\Cref{tab:act_obs_spaces}). Each policy is instantiated as a neural network, and the output of the neural network is used as input to a low-level skill.

The \textsc{Ball Duel} policy, trained in a 2 vs.\ 0 AbstractSim environment, develops ball control skills through velocity-based maneuvering. During training, the policy is rewarded for moving toward the ball, moving the ball toward the goal and scoring.  Despite the absence of opponents in training, its proficiency in ball handling makes it effective in real-world contested situations. However, we identified three respects in which this policy underperformed due to the sim2real gap: slow movement when far from the ball, imprecise kicking, and struggles in near-goal situations. 

The \textsc{Mid-field} policy addresses the \textsc{Ball Duel} policy's limitations in walking and kicking. Developed in a 1 vs.\ 0 AbstractSim environment, it uses the B-Human robot architecture's walk-and-kick skill. During training it is rewarded for moving the ball toward the goal and scoring. The \textsc{Mid-field} policy outputs a kick angle that parameterizes a low-level walk-and-kick skill previously developed by the B-Human team. The walk-and-kick skill incorporates a path planner for obstacle avoidance. By employing a different lower-level skill, the \textsc{Mid-field} policy sacrifices precise velocity control in favor of enhanced movement speed and kicking accuracy and excels in less contested scenarios.

The \textsc{Near-goal} policy is designed for critical situations where the ball is close to the goal, often requiring decisive and precise movement to score. To achieve the necessary precision, we trained this policy using a 1 vs.\ 0 scenario in the high-fidelity SimRobot simulator. During training, the agent was given a positive reward for scoring and a negative reward for moving the ball too far from the goal along with shaping rewards to encourage learning. This approach revealed subtle yet effective strategies; for instance, the \textsc{Near-goal} policy learned to make small lateral movements to effectively bump the ball towards the goal, proving more efficient than actively kicking.

Finally, the \textsc{Positioning} policy guides the robot's movement when a teammate is closer to the ball. It considers both the ball's position and a manually defined strategy position. This policy was rewarded for moving toward its predefined strategy position, keeping the ball in view and avoiding opponents. 

\subsection{Heuristic Policy Selection}
\label{sec:sub_heuristic}
Our system employs heuristic-based selection to dynamically select from among the four specialized sub-behavior policies based on specific game situations. The \textsc{Positioning} policy is activated when a teammate is estimated to be closer to the ball and upright, guiding the agent to supportive field positions. The \textsc{Near-goal} policy, trained in high-fidelity SimRobot, takes over when the agent is near the ball within the opposing goal box. The \textsc{Ball Duel} policy is engaged when an opponent robot is within half a meter of the ball, managing contested situations with precise ball control. The \textsc{Mid-field} policy serves as the default, enabling efficient field navigation and accurate kicking when no other conditions are met. 

This heuristic approach enables dynamic playstyle adjustments and integration of new policies, enhancing our team's adaptability to various game scenarios. For instance, after observing the \textsc{Near-goal} sub-behavior's effective performance, we expanded its activation region. Our system's flexibility allows for updates to existing policies and integration of new ones between matches, facilitating continuous improvement. 
This adaptability proved crucial in our performance evolution throughout the competition, enabling us to turn a close 2-1 victory in our first game to a resounding 8-0 victory in our last against the same team.

\section{EMPIRICAL ANALYSIS}
In this section, we study the key decisions that led to our first-place finish in the RoboCup competition. We focus on three elements that we hypothesized contributed to our success: heuristic policy selection, training policies in different simulation fidelities, and utilizing distinct action spaces for the \textsc{Ball Duel} and \textsc{Mid-field} policies. We conduct experiments on physical robots and in high-fidelity simulation (SimRobot). % to empirically assess these decisions and verify their impact on real-world performance.

\subsection{Heuristic Policy Conditioning}

\begin{figure}[ht]
\small
    \centering
    \arrayrulecolor{gray!80} % Set the color of the table lines
    \rowcolors{2}{white}{gray!25} % Start from the second row
    \begin{tabularx}{\linewidth}{|X|X|}
    \hline
    Experiment & \makecell[l]{Physical Successes} \\
    \hline
    Full Suite & $\mathbf{6/10 \pm 3}$ \\
    No \textsc{Mid-field} & $0/10 \pm 0$ \\
    No \textsc{Near-goal} & $4/10 \pm 3$ \\
    No \textsc{Ball Duel} & $3/10 \pm 3$ \\
    \hline
        
    \end{tabularx}
    \caption{Evaluation of policy decomposition on success rate against a defender robot. Success is a goal, failure is an out of bounds or timeout of a minute. Higher is better. Confidence intervals are 95\% bootstrapped.}
    \label{fig:feature1results}

    \vspace{-3.5mm}
\end{figure}

The first experiment evaluates the effectiveness of our policy decomposition and heuristic-based policy selection. We test each policy's performance on physical robots against a weakened defender and goalie\footnote{The goalie code in our system is manually defined, as the behavior for this role is relatively simple to implement.} with disabled kicking abilities. The setup starts the attacker agent with possession of the ball in a 1 vs.\ 2 scoring evaluation. The results, presented in \Cref{fig:feature1results}, show that the full suite of policies outperforms systems where one policy is removed, indicating that each policy plays a crucial role in the overall performance of the system.

% In simulation, we analyze the same ablations with one policy removed but record the win rate over a large number of games. In this setting we see... TODO: Waiting on results.

\subsection{Simulation Fidelity}

\begin{figure}[ht]
\vspace{2mm}
    \centering
    % Subfigure for the table
    \begin{subfigure}[b]{\linewidth}
    \small
        \centering
        \arrayrulecolor{gray!80} % Set the color of the table lines
        \begin{tabularx}{\linewidth}{|c|XXX|}
        \hline
        Experiment & \makecell[l]{Training\\Simulation} & \makecell[l]{Physical\\Success} & \makecell[l]{Simulation\\Success} \\
        \hline
        \multirow{2}{*}{\makecell[c]{Goalie}} 
        & \cellcolor{gray!25}AbstractSim & \cellcolor{gray!25}$7/10 \pm 3$ & \cellcolor{gray!25}$\mathbf{77/100 \pm 8}$ \\
        & SimRobot & $\mathbf{9/10 \pm 1.5}$ & $62/100 \pm 9$ \\
        \hline
        \multirow{2}{*}{\makecell[c]{Goalie and\\Defender}} 
        & \cellcolor{gray!25}AbstractSim & \cellcolor{gray!25}$4/10 \pm 3$ & \cellcolor{gray!25}$\mathbf{62/100 \pm 9}$ \\
        & SimRobot & $\mathbf{9/10 \pm 1.5}$ & $60/100 \pm 10$ \\
        \hline
        \end{tabularx}
        \caption{Simulation type success results. Higher is better.}
        \label{fig:training_sim}
    \end{subfigure}
    
    % Subfigure for the image
    \begin{subfigure}[b]{\linewidth}
        \centering
        \includegraphics[width=\linewidth]{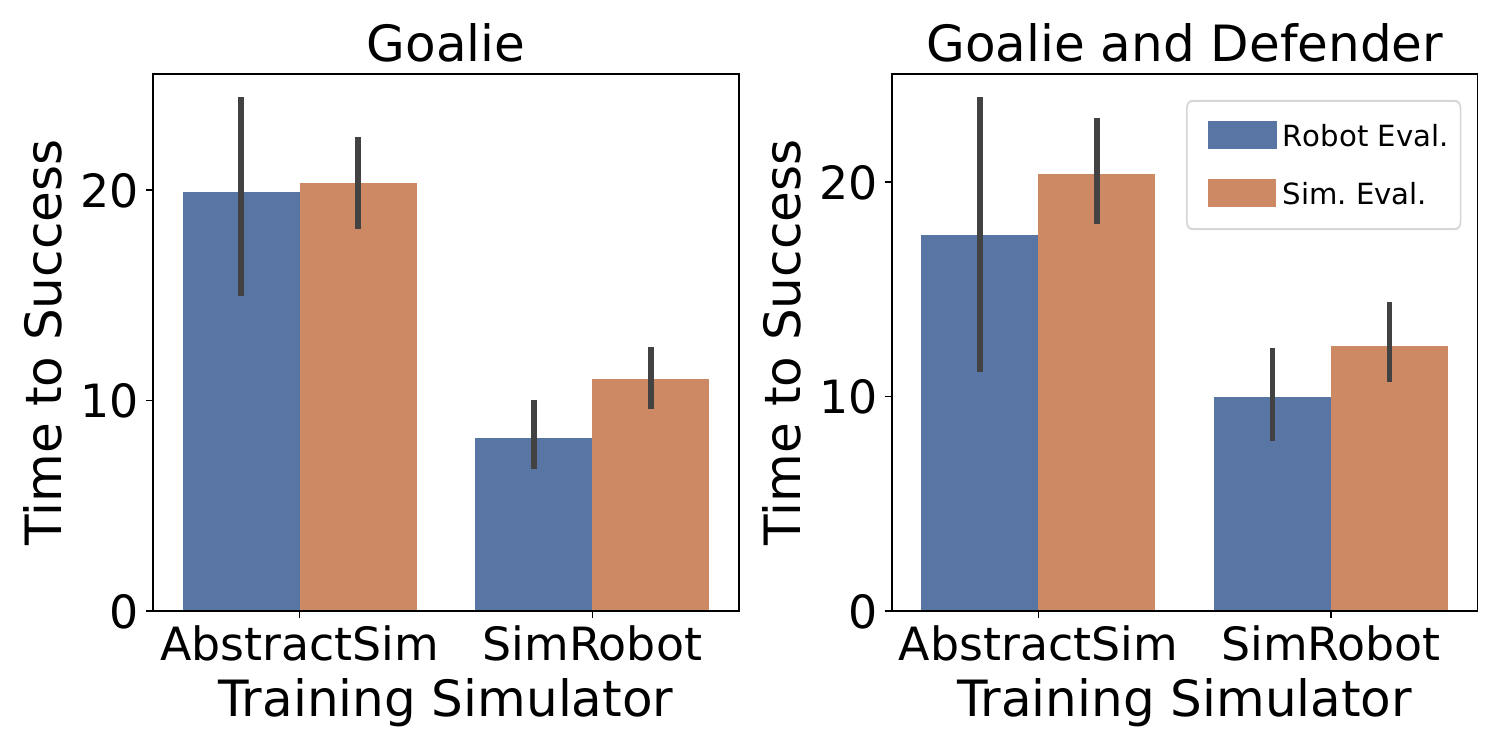}
        \caption{Simulation type time to success results. Lower is better.}
    \end{subfigure}
    
    \caption{Results from training simulation fidelity experiments. We compare low-fidelity AbstractSim trained policies to high-fidelity SimRobot trained policies. We test against a setup with only a goalie and against a setup with a goalie and defender. Success is a goal. Failure is a timeout of a minute or out of bounds. Confidence intervals are 95\% bootstrap confidence intervals.}
    \label{fig:feature2results}
\end{figure}

The second experiment examines the impact of simulation fidelity on policy performance. Specifically, we focus on comparing the effectiveness of training the \textsc{Near-goal} policy in high-fidelity SimRobot versus low-fidelity AbstractSim. We trained policies to convergence, with initialization within the goal box. We then tested these policies in two scenarios: a goalie only and a defender and goalie together. In this evaluation we also perform a scoring test with the evaluated policy given possession of the ball to start. The results, shown in \Cref{fig:feature2results}, demonstrate that on the physical robots, the SimRobot-trained policy achieves a significantly higher success rate and shorter time to score, showing the performance boost gained by using SimRobot-trained policies. Interestingly, in the simulation experiments, the AbstractSim-trained policies outperform the SimRobot-trained policy. The experiment results indicate that the AbstractSim-trained policy performs well in simulations but fails to generalize to the real world.

\subsection{Action Spaces}

\begin{figure}[ht]
\vspace{2mm}
    \centering
    % Subfigure for the table
    \begin{subfigure}[b]{\linewidth}
    \small
        \centering
        \arrayrulecolor{gray!80} % Set the color of the table lines
        \rowcolors{2}{white}{gray!25} % Start from the second row
        \begin{tabularx}{\linewidth}{|X|XX|}
        \hline
        Experiment & \makecell[l]{Physical\\Success} & \makecell[l]{Simulation\\Success} \\
        \hline
        \makecell[l]{Walk at\\Relative Speed} & $\mathbf{7/10 \pm 3}$ & $\mathbf{41/100 \pm 15}$ \\
        Walk to Point & $1/10 \pm 1.5$ & $11/100 \pm 6$ \\
        \hline
        \end{tabularx}
        \caption{Evaluation of action spaces. Success moving the ball past the opponent with control. Failure is a timeout at a minute or losing control of the ball. Higher is better.}
        \label{fig:policy_decomp}
    \end{subfigure}
    
    % Subfigure for the image
    \begin{subfigure}[b]{\linewidth}
        \centering
        \includegraphics[width=\linewidth]{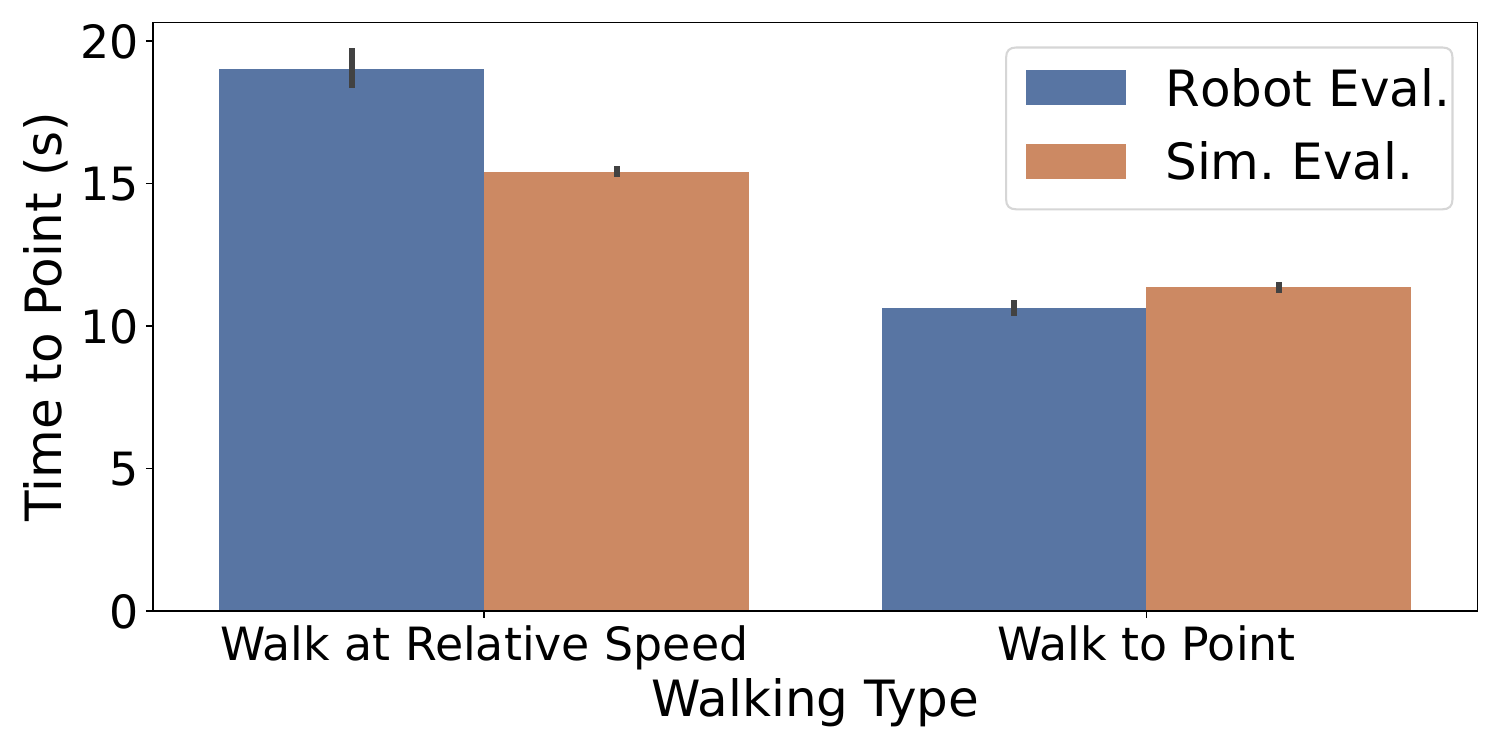}
        \caption{Walking Type Experimental Results. Time to reach a point on the opposite side of the field as the robot. Lower is better.}
        \label{fig:feature3results}
    \end{subfigure}
    
    \caption{Results from action space experiments. In \Cref{fig:policy_decomp} we show the success of dribbling around an opponent. In \Cref{fig:feature3results} we show the time to walk to a point 4m away from the robot. Confidence intervals are 95\% bootstrap confidence intervals.}
    \label{fig:combined_results}
    \vspace{-3mm}
\end{figure}

The third experiment examines the trade-offs between the two action spaces used in our \textsc{Ball Duel} and \textsc{Mid-Field} policies. We conduct two tests to evaluate these conditions and the results are displayed in \Cref{fig:combined_results}.

The first test assesses the agent’s ability to move the ball around an opposing robot. Here, we use our defender code with kicking disabled as the opposing robot and the attacking agent is started with the ball. In this scenario, the walk-at-relative-speed action space has a significantly higher success rate than the walk-to-point action space. This is due to the limitations of the walk-to-point action space in precise ball manipulation. In contrast, the dribble policy has precise movement control.

The second test measures the time it takes for the robot to walk to a point 4m away. A lower time indicates faster walking. Qualitatively, the walk-to-point action space had smoother movement because it has a more stable desired location. In contrast, the walk-at-relative-speed action space adjusts the desired velocity at every timestep, resulting in slower movement.

\section{DISCUSSION AND LIMITATIONS}

Our SPL case study offers lessons for similar domains. Decomposing complex RL tasks into learnable sub-behaviors allows faster training and facilitates adjustments to the overall behavior post-training. Bootstrapping off of existing classical robotics stacks can also make RL more feasible with limited resources.
Our approach also shows that matching simulator fidelity to the target task is crucial. For tasks requiring both global coverage and local precision, using multiple fidelities of simulation can enhance overall performance.

As an example real-world application where our lessons could be applied, we consider a disaster response scenario.
% In a multi-team disaster response scenario, diverse robotic systems could benefit from our findings. 
% 
Response teams with robots could use simplified simulators to develop general exploration policies, while utilizing high-fidelity simulations to refine task-specific sub-behaviors like debris removal or medical assessment. These sub-behaviors can be integrated together with the heuristic-based sub-behavior selection scheme. By combining existing modules for perception and low-level control with RL-trained high-level decision-making, teams can reduce the computational burden compared to end-to-end training. This framework allows for rapid deployment and on-site fine-tuning of robot behaviors without full retraining, thereby enhancing the efficiency of joint rescue operations.
% potential application domain, elaboraed
% Consider a multi-team disaster response scenario using diverse robotic systems. Each team deploys specialized robots, all navigating a common search area. This scenario could benefit from our findings through behavior decomposition, multi-fidelity simulation, and a hybrid approach. Teams could develop a general exploration policy using a simplified simulator, then implement specialized sub-behaviors for each robot type. A low-fidelity simulator could train broad navigation, while high-fidelity simulations refine task-specific actions like debris removal or medical assessment. Reusing robust classical modules such as locomotion controllers alongside RL-trained behaviors for high-level decision-making eliminates the need to train locomotion. This framework enables rapid deployment and on-site fine-tuning of robot behaviors without full retraining, potentially improving the efficiency and effectiveness of joint rescue operations.

%Limitation, Compressed
Our current approach faces several limitations that future work could address. First, we aim to develop multi-agent training methods for complex team behaviors. Currently, we rely on hand-coded sub-policy decomposition and training scenarios, which is effective but potentially leaves room for improvement. 
Second, our heuristic approach also does not take into account other agents sub-policy selection. As each agent individually computes the role they are playing, the role can change rapidly. Communication and a bidding system is a future direction to implement for our role switching heuristic.
Other future work can explore learning of sub-behavior selection, investigate methods for balancing high and low-fidelity simulators, or explore human-in-the-loop methods to further leverage expert knowledge in decision-making and strategy control.

\section{CONCLUSION}

    Robot soccer and the annual RoboCup competition is a research challenge task designed to spur innovation in building complete robot architectures that can operate in dynamic, partially observable, and adversarial domains.
    In this paper, we have described an RL approach for developing high-level behaviors for the NAO robot that won the Challenge Shield division of the 2024 RoboCup Standard Platform League competition.
    This work provides insights and lessons for using model-free RL as a primary driver of decision-making in dynamic, multi-agent and partially observable robot tasks where end-to-end RL may be intractable yet domain complexity suggests that manual programming of behaviors is likely suboptimal.
    In addition to describing our system, we conducted empirical analysis of three critical components: heuristic-based policy selection, varying simulation fidelity and different action spaces.
    The results of this analysis provide further lessons for the application of RL in domains with similar challenges.
    This work demonstrates the promise of RL for developing robot behaviors in complex, dynamic, partially observable, and multi-agent domains.

%%%%%%%%%%%%%%%%%%%%




\section*{ACKNOWLEDGMENT}
A portion of this work has taken place in the Learning Agents Research
Group (LARG) at UT Austin. 
LARG research is supported in part by NSF
(FAIN-2019844, NRT-2125858), ONR (N00014-18-2243), ARO
(W911NF-23-2-0004, W911NF-17-2-0181), DARPA (Cooperative Agreement
HR00112520004 on Ad Hoc Teamwork), Lockheed Martin, and UT Austin’s
Good Systems grand challenge. Peter Stone serves as the Executive
Director of Sony AI America and receives financial compensation for
this work. The terms of this arrangement have been reviewed and
approved by the University of Texas at Austin in accordance with its
policy on objectivity in research.
Josiah Hanna acknowledges support from NSF (IIS-2410981), American Family Insurance through a research partnership with the University of Wisconsin—Madison’s Data Science Institute, the Wisconsin Alumni Research Foundation, and Sandia National Labs through a University Partnership Award.

\clearpage

\bibliographystyle{IEEEtran}
\bibliography{refs}

\begin{thebibliography}{10}
\providecommand{\url}[1]{#1}
\csname url@samestyle\endcsname
\providecommand{\newblock}{\relax}
\providecommand{\bibinfo}[2]{#2}
\providecommand{\BIBentrySTDinterwordspacing}{\spaceskip=0pt\relax}
\providecommand{\BIBentryALTinterwordstretchfactor}{4}
\providecommand{\BIBentryALTinterwordspacing}{\spaceskip=\fontdimen2\font plus
\BIBentryALTinterwordstretchfactor\fontdimen3\font minus
  \fontdimen4\font\relax}
\providecommand{\BIBforeignlanguage}[2]{{%
\expandafter\ifx\csname l@#1\endcsname\relax
\typeout{** WARNING: IEEEtran.bst: No hyphenation pattern has been}%
\typeout{** loaded for the language `#1'. Using the pattern for}%
\typeout{** the default language instead.}%
\else
\language=\csname l@#1\endcsname
\fi
#2}}
\providecommand{\BIBdecl}{\relax}
\BIBdecl

\bibitem{akkaya2019solving}
I.~Akkaya, M.~Andrychowicz, M.~Chociej, M.~Litwin, B.~McGrew, A.~Petron,
  A.~Paino, M.~Plappert, G.~Powell, R.~Ribas \emph{et~al.}, ``Solving rubik's
  cube with a robot hand,'' \emph{arXiv preprint arXiv:1910.07113}, 2019.

\bibitem{li2021reinforcement}
Z.~Li, X.~Cheng, X.~B. Peng, P.~Abbeel, S.~Levine, G.~Berseth, and K.~Sreenath,
  ``Reinforcement learning for robust parameterized locomotion control of
  bipedal robots,'' in \emph{2021 IEEE International Conference on Robotics and
  Automation (ICRA)}.\hskip 1em plus 0.5em minus 0.4em\relax IEEE, 2021, pp.
  2811--2817.

\bibitem{dambrosio2024achievinghumanlevelcompetitive}
\BIBentryALTinterwordspacing
D.~B. D'Ambrosio, S.~Abeyruwan, L.~Graesser, A.~Iscen, H.~B. Amor, A.~Bewley,
  B.~J. Reed, K.~Reymann, L.~Takayama, Y.~Tassa, K.~Choromanski, E.~Coumans,
  D.~Jain, N.~Jaitly, N.~Jaques, S.~Kataoka, Y.~Kuang, N.~Lazic, R.~Mahjourian,
  S.~Moore, K.~Oslund, A.~Shankar, V.~Sindhwani, V.~Vanhoucke, G.~Vesom, P.~Xu,
  and P.~R. Sanketi, ``Achieving human level competitive robot table tennis,''
  2024. [Online]. Available: \url{https://arxiv.org/abs/2408.03906}
\BIBentrySTDinterwordspacing

\bibitem{nardi2014robocup}
D.~Nardi, I.~Noda, F.~Ribeiro, P.~Stone, O.~von Stryk, and M.~Veloso, ``Robocup
  soccer leagues,'' \emph{AI Magazine}, vol.~35, no.~3, pp. 77--85, 2014.

\bibitem{kitano1997robocup}
H.~Kitano, M.~Asada, Y.~Kuniyoshi, I.~Noda, and E.~Osawa, ``Robocup: The robot
  world cup initiative,'' in \emph{Proceedings of the first international
  conference on Autonomous agents}, 1997, pp. 340--347.

\bibitem{tang2024deep}
C.~Tang, B.~Abbatematteo, J.~Hu, R.~Chandra, R.~Mart{\'\i}n-Mart{\'\i}n, and
  P.~Stone, ``Deep reinforcement learning for robotics: A survey of real-world
  successes,'' \emph{arXiv preprint arXiv:2408.03539}, 2024.

\bibitem{siekmann2020learning}
J.~Siekmann, S.~Valluri, J.~Dao, L.~Bermillo, H.~Duan, A.~Fern, and J.~Hurst,
  ``Learning memory-based control for human-scale bipedal locomotion,''
  \emph{arXiv preprint arXiv:2006.02402}, 2020.

\bibitem{castillo2022reinforcement}
G.~A. Castillo, B.~Weng, W.~Zhang, and A.~Hereid, ``Reinforcement
  learning-based cascade motion policy design for robust 3d bipedal
  locomotion,'' \emph{IEEE Access}, vol.~10, pp. 20\,135--20\,148, 2022.

\bibitem{duan2024learning}
H.~Duan, B.~Pandit, M.~S. Gadde, B.~Van~Marum, J.~Dao, C.~Kim, and A.~Fern,
  ``Learning vision-based bipedal locomotion for challenging terrain,'' in
  \emph{2024 IEEE International Conference on Robotics and Automation
  (ICRA)}.\hskip 1em plus 0.5em minus 0.4em\relax IEEE, 2024, pp. 56--62.

\bibitem{beranek2021behavior}
R.~Beranek, M.~Karimi, and M.~Ahmadi, ``A behavior-based reinforcement learning
  approach to control walking bipedal robots under unknown disturbances,''
  \emph{IEEE/ASME Transactions on Mechatronics}, vol.~27, no.~5, pp.
  2710--2720, 2021.

\bibitem{kouppas2021hybrid}
C.~Kouppas, M.~Saada, Q.~Meng, M.~King, and D.~Majoe, ``Hybrid autonomous
  controller for bipedal robot balance with deep reinforcement learning and
  pattern generators,'' \emph{Robotics and Autonomous Systems}, vol. 146, p.
  103891, 2021.

\bibitem{qin2024heuristics}
D.~Qin, G.~Zhang, Z.~Zhu, T.~Chen, W.~Zhu, X.~Rong, A.~Xie, and Y.~Li, ``A
  heuristics-based reinforcement learning method to control bipedal robots,''
  \emph{Int. J. Humanoid Robot}, 2024.

\bibitem{li2019using}
T.~Li, H.~Geyer, C.~G. Atkeson, and A.~Rai, ``Using deep reinforcement learning
  to learn high-level policies on the atrias biped,'' in \emph{2019
  International Conference on Robotics and Automation (ICRA)}.\hskip 1em plus
  0.5em minus 0.4em\relax IEEE, 2019, pp. 263--269.

\bibitem{nachum2019multi}
O.~Nachum, M.~Ahn, H.~Ponte, S.~Gu, and V.~Kumar, ``Multi-agent manipulation
  via locomotion using hierarchical sim2real,'' \emph{arXiv preprint
  arXiv:1908.05224}, 2019.

\bibitem{li2020learning}
T.~Li, N.~Lambert, R.~Calandra, F.~Meier, and A.~Rai, ``Learning generalizable
  locomotion skills with hierarchical reinforcement learning,'' in \emph{2020
  IEEE International Conference on Robotics and Automation (ICRA)}.\hskip 1em
  plus 0.5em minus 0.4em\relax IEEE, 2020, pp. 413--419.

\bibitem{li2021planning}
T.~Li, R.~Calandra, D.~Pathak, Y.~Tian, F.~Meier, and A.~Rai, ``Planning in
  learned latent action spaces for generalizable legged locomotion,''
  \emph{IEEE Robotics and Automation Letters}, vol.~6, no.~2, pp. 2682--2689,
  2021.

\bibitem{truong2023rethinking}
J.~Truong, M.~Rudolph, N.~H. Yokoyama, S.~Chernova, D.~Batra, and A.~Rai,
  ``Rethinking sim2real: Lower fidelity simulation leads to higher sim2real
  transfer in navigation,'' in \emph{Conference on Robot Learning}.\hskip 1em
  plus 0.5em minus 0.4em\relax PMLR, 2023, pp. 859--870.

\bibitem{zhang2024back}
Y.~Zhang, Y.~Hu, Y.~Song, D.~Zou, and W.~Lin, ``Back to newton's laws: Learning
  vision-based agile flight via differentiable physics,'' \emph{arXiv preprint
  arXiv:2407.10648}, 2024.

\bibitem{bhola2023multi}
S.~Bhola, S.~Pawar, P.~Balaprakash, and R.~Maulik, ``Multi-fidelity
  reinforcement learning framework for shape optimization,'' \emph{Journal of
  Computational Physics}, vol. 482, p. 112018, 2023.

\bibitem{cutler2014reinforcement}
M.~Cutler, T.~J. Walsh, and J.~P. How, ``Reinforcement learning with
  multi-fidelity simulators,'' in \emph{2014 IEEE International Conference on
  Robotics and Automation (ICRA)}, 2014, pp. 3888--3895.

\bibitem{cutler2015real}
------, ``Real-world reinforcement learning via multifidelity simulators,''
  \emph{IEEE Transactions on Robotics}, vol.~31, no.~3, pp. 655--671, 2015.

\bibitem{khairy2024multi}
S.~Khairy and P.~Balaprakash, ``Multi-fidelity reinforcement learning with
  control variates,'' \emph{Neurocomputing}, p. 127963, 2024.

\bibitem{beard2022black}
J.~J. Beard and A.~Baheri, ``Black-box safety validation of autonomous systems:
  A multi-fidelity reinforcement learning approach,'' \emph{arXiv preprint
  arXiv:2203.03451}, 2022.

\bibitem{suryan2020multifidelity}
V.~Suryan, N.~Gondhalekar, and P.~Tokekar, ``Multifidelity reinforcement
  learning with gaussian processes: model-based and model-free algorithms,''
  \emph{IEEE Robotics \& Automation Magazine}, vol.~27, no.~2, pp. 117--128,
  2020.

\bibitem{ryou2024multi}
G.~Ryou, G.~Wang, and S.~Karaman, ``Multi-fidelity reinforcement learning for
  time-optimal quadrotor re-planning,'' \emph{arXiv preprint arXiv:2403.08152},
  2024.

\bibitem{hong2021ai}
C.~Hong, I.~Jeong, L.~F. Vecchietti, D.~Har, and J.-H. Kim, ``Ai world cup:
  robot-soccer-based competitions,'' \emph{IEEE Transactions on Games},
  vol.~13, no.~4, pp. 330--341, 2021.

\bibitem{smit2023scaling}
A.~Smit, H.~A. Engelbrecht, W.~Brink, and A.~Pretorius, ``Scaling multi-agent
  reinforcement learning to full 11 versus 11 simulated robotic football,''
  \emph{Autonomous Agents and Multi-Agent Systems}, vol.~37, no.~1, p.~20,
  2023.

\bibitem{antonioni2021game}
E.~Antonioni, V.~Suriani, F.~Riccio, and D.~Nardi, ``Game strategies for
  physical robot soccer players: a survey,'' \emph{IEEE Transactions on Games},
  vol.~13, no.~4, pp. 342--357, 2021.

\bibitem{AB05}
P.~Stone, R.~S. Sutton, and G.~Kuhlmann, ``Reinforcement learning for
  {R}obo{C}up-soccer keepaway,'' \emph{Adaptive Behavior}, vol.~13, no.~3, pp.
  165--188, 2005.

\bibitem{abreu2023designingskilledsoccerteam}
\BIBentryALTinterwordspacing
M.~Abreu, L.~P. Reis, and N.~Lau, ``Designing a skilled soccer team for
  robocup: Exploring skill-set-primitives through reinforcement learning,''
  2023. [Online]. Available: \url{https://arxiv.org/abs/2312.14360}
\BIBentrySTDinterwordspacing

\bibitem{huang2021tikickplayingmultiagentfootball}
\BIBentryALTinterwordspacing
S.~Huang, W.~Chen, L.~Zhang, S.~Xu, Z.~Li, F.~Zhu, D.~Ye, T.~Chen, and J.~Zhu,
  ``Tikick: Towards playing multi-agent football full games from single-agent
  demonstrations,'' 2021. [Online]. Available:
  \url{https://arxiv.org/abs/2110.04507}
\BIBentrySTDinterwordspacing

\bibitem{lin2023tizeromasteringmultiagentfootball}
\BIBentryALTinterwordspacing
F.~Lin, S.~Huang, T.~Pearce, W.~Chen, and W.-W. Tu, ``Tizero: Mastering
  multi-agent football with curriculum learning and self-play,'' 2023.
  [Online]. Available: \url{https://arxiv.org/abs/2302.07515}
\BIBentrySTDinterwordspacing

\bibitem{liu2022motor}
S.~Liu, G.~Lever, Z.~Wang, J.~Merel, S.~A. Eslami, D.~Hennes, W.~M. Czarnecki,
  Y.~Tassa, S.~Omidshafiei, A.~Abdolmaleki \emph{et~al.}, ``From motor control
  to team play in simulated humanoid football,'' \emph{Science Robotics},
  vol.~7, no.~69, p. eabo0235, 2022.

\bibitem{liu2019emergent}
S.~Liu, G.~Lever, J.~Merel, S.~Tunyasuvunakool, N.~Heess, and T.~Graepel,
  ``Emergent coordination through competition,'' \emph{arXiv preprint
  arXiv:1902.07151}, 2019.

\bibitem{da2021deep}
I.~J. da~Silva, D.~H. Perico, T.~P.~D. Homem, and R.~A. da~Costa~Bianchi,
  ``Deep reinforcement learning for a humanoid robot soccer player,''
  \emph{Journal of Intelligent \& Robotic Systems}, vol. 102, no.~3, p.~69,
  2021.

\bibitem{merke2002karlsruhe}
A.~Merke and M.~Riedmiller, ``Karlsruhe brainstormers-a reinforcement learning
  approach to robotic soccer,'' in \emph{RoboCup 2001: Robot Soccer World Cup V
  5}.\hskip 1em plus 0.5em minus 0.4em\relax Springer, 2002, pp. 435--440.

\bibitem{riedmiller2009reinforcement}
M.~Riedmiller, T.~Gabel, R.~Hafner, and S.~Lange, ``Reinforcement learning for
  robot soccer,'' \emph{Autonomous Robots}, vol.~27, pp. 55--73, 2009.

\bibitem{huang2023creating}
X.~Huang, Z.~Li, Y.~Xiang, Y.~Ni, Y.~Chi, Y.~Li, L.~Yang, X.~B. Peng, and
  K.~Sreenath, ``Creating a dynamic quadrupedal robotic goalkeeper with
  reinforcement learning,'' in \emph{2023 IEEE/RSJ International Conference on
  Intelligent Robots and Systems (IROS)}.\hskip 1em plus 0.5em minus
  0.4em\relax IEEE, 2023, pp. 2715--2722.

\bibitem{ji2023dribblebot}
Y.~Ji, G.~B. Margolis, and P.~Agrawal, ``Dribblebot: Dynamic legged
  manipulation in the wild,'' in \emph{2023 IEEE International Conference on
  Robotics and Automation (ICRA)}.\hskip 1em plus 0.5em minus 0.4em\relax IEEE,
  2023, pp. 5155--5162.

\bibitem{ji2022hierarchical}
Y.~Ji, Z.~Li, Y.~Sun, X.~B. Peng, S.~Levine, G.~Berseth, and K.~Sreenath,
  ``Hierarchical reinforcement learning for precise soccer shooting skills
  using a quadrupedal robot,'' in \emph{2022 IEEE/RSJ International Conference
  on Intelligent Robots and Systems (IROS)}.\hskip 1em plus 0.5em minus
  0.4em\relax IEEE, 2022, pp. 1479--1486.

\bibitem{haarnoja2024learning}
T.~Haarnoja, B.~Moran, G.~Lever, S.~H. Huang, D.~Tirumala, J.~Humplik,
  M.~Wulfmeier, S.~Tunyasuvunakool, N.~Y. Siegel, R.~Hafner \emph{et~al.},
  ``Learning agile soccer skills for a bipedal robot with deep reinforcement
  learning,'' \emph{Science Robotics}, vol.~9, no.~89, p. eadi8022, 2024.

\bibitem{ros2009case}
R.~Ros, J.~L. Arcos, R.~L. De~Mantaras, and M.~Veloso, ``A case-based approach
  for coordinated action selection in robot soccer,'' \emph{Artificial
  intelligence}, vol. 173, no. 9-10, pp. 1014--1039, 2009.

\bibitem{BHumanCodeReleaseDocumentation2023}
T.~R{\"o}fer, T.~Laue, F.~B{\"o}se, A.~Hasselbring, J.~Lienhoop, L.~M.
  Monnerjahn, P.~Reichenberg, and S.~Schreiber, ``{B}-{H}uman code release
  documentation 2023,'' 2023, only available online:
  \url{https://docs.b-human.de/coderelease2023/}.

\bibitem{tirumala2024learning}
D.~Tirumala, M.~Wulfmeier, B.~Moran, S.~Huang, J.~Humplik, G.~Lever,
  T.~Haarnoja, L.~Hasenclever, A.~Byravan, N.~Batchelor \emph{et~al.},
  ``Learning robot soccer from egocentric vision with deep reinforcement
  learning,'' \emph{arXiv preprint arXiv:2405.02425}, 2024.

\bibitem{schulman2017proximal}
J.~Schulman, F.~Wolski, P.~Dhariwal, A.~Radford, and O.~Klimov, ``Proximal
  policy optimization algorithms,'' \emph{arXiv preprint arXiv:1707.06347},
  2017.

\bibitem{stable-baselines3}
\BIBentryALTinterwordspacing
A.~Raffin, A.~Hill, A.~Gleave, A.~Kanervisto, M.~Ernestus, and N.~Dormann,
  ``Stable-baselines3: Reliable reinforcement learning implementations,''
  \emph{Journal of Machine Learning Research}, vol.~22, no. 268, pp. 1--8,
  2021. [Online]. Available: \url{http://jmlr.org/papers/v22/20-1364.html}
\BIBentrySTDinterwordspacing

\end{thebibliography}

% \begin{thebibliography}{99}

% \end{thebibliography}

\end{document}